\documentclass[runningheads]{llncs}
\usepackage[T1]{fontenc}
\usepackage{graphicx}
\usepackage{amsmath}
\usepackage[numbers,sort&compress]{natbib}
\usepackage{booktabs}
\usepackage{makecell}
\usepackage[svgnames,table]{xcolor}

\renewcommand{\bf}[1]{\cellcolor{black!20}\textbf{#1}}

\begin{document}
\title{Volatility forecasting using Deep Learning and sentiment analysis}
%
%
\author{V Ncume\inst{1} \and
T. L van Zyl\inst{2}\orcidID{0000-0003-4281-630X} \and
A Paskaramoorthy \inst{3}\orcidID{0000-0002-7812-5909}}

\authorrunning{V Ncume \emph{et al.}}

\institute{Computer Science and Applied Mathematics, University of the Witwatersrand, Johannesburg, South Africa \\
\email{vuyoncume68@gmail.com}
\and
Institute for Intelligent Systems, University of Johannesburg, Johannesburg, South Africa \\
\email{tvanzyl@gmail.com}
\and
Department of Statistical Sciences, University of Cape Town, Cape Town, South Africa \\
\email{ab.paskaramoorthy@gmail.com}}

\maketitle

\begin{abstract}
Several studies have shown that deep learning models can provide more accurate volatility forecasts than the traditional methods used within this domain. This paper presents a composite model that merges a deep learning approach with sentiment analysis for predicting market volatility. To classify public sentiment, we use a Convolutional Neural Network, which obtained data from Reddit global news headlines. We then describe a composite forecasting model, a Long-Short-Term-Memory Neural Network method, to use historical sentiment and the previous day's volatility to make forecasts. We employed this method on the past volatility of the S\&P500 and the major BRICS indices to corroborate its effectiveness. Our results demonstrate that including sentiment can improve Deep Learning volatility forecasting models. However, in contrast to return forecasting, the performance benefits of including sentiment for volatility forecasting appears to be market specific.

\end{abstract}

\section{Introduction}
Deep Learning has shown to be useful in sequential data prediction tasks such as time series forecasting and text prediction. Given sufficient compute power and time, Deep Learning algorithms are able to learn from large datasets and outperform traditional machine learning and statistical techniques. Consequently, there is increasing interest in using Deep Learning for economic and financial forecasting owing to its successes in other domains. A growing literature investigates whether Deep Learning algorithms with various architectures can be used to make predictions in financial markets that can be exploited for profit \cite{jing2021hybrid, mathonsi2020prediction, mathonsi2022multivariate, laher2021deep}.

Previous work in the financial time series forecasting domain has acknowledged the importance of sentiment in predicting financial markets, and thus we seek to use sentiment data in conjunction with a deep learning model to increase prediction accuracy. Text from the internet is increasingly becoming more relevant as an important type of data to be included in predictive models. For example, \cite{jin2013forex} develops a prediction model that combines news events and financial data to predict the fluctuation of foreign currency. \cite{chen2014wisdom} shows that opinions on popular online platforms are strong predictors of earnings surprises and future market returns for stocks. Other studies have corroborated that social media posts are useful for prediction in finance (for example, \cite{yu2013impact} and \cite{wang2018combining}). 

Deep learning techniques have been used to forecast market returns in various ways and have shown to be more accurate at making predictions when sentiment is included as an input.  For example, In \cite{mehtab2019robust} and \cite{muthivhi2022fusion}, a Long Short Term Memory Neural Network (LSTM) is used to forecast the stock closing price along with data from Twitter to gauge public sentiment. \cite{jing2021hybrid} proposes a hybrid algorithm where a CNN is used for classifying sentiments, which were used as inputs into an LSTM Neural Network to predict stock prices with similar results to \cite{mehtab2019robust} and \cite{muthivhi2022fusion}. 

Whilst forecasting market returns is receiving increased attention, using Deep Learning models for volatility forecasting (another important problem in finance) has been largely unexplored. Volatility forecasting can be seen as easier than return forecasting due to the presence of second-order autocorrelation in empirical returns (known as ``volatility clustering``). Volatility is typically modeled using traditional time series models, such as the Generalized Autoregressive Conditional Heteroskedasticity (GARCH) model or its extensions \cite{andersen1998answering}. However, studies by \cite{liu2019novel}, \cite{ge2022neural}, \cite{petrozziello2022deep} and \cite{xiong2015deep} show that Deep Learning methods can outperform the more traditional methods in the volatility forecasting domain. 

The study by \cite{ge2022neural} shows that there is still a large gap between the state-of-the-art deep learning techniques available and their use in the volatility forecasting domain. We look to close this gap by proposing a hybrid deep learning model which forecasts sentiment, which is used in turn to forecast the volatility of a market index. More specifically, we combined a Convolutional Neural Network (CNN) for sentiment analysis and a Long Term Short Term Memory~\cite{freeborough2022investigating} Neural Network for the volatility predictions. We used this hybrid approach to forecast the past volatility of the S\&P500 and the major BRICS indices~\cite{cawood2022evaluating}. Our approach is similar to \cite{jing2021hybrid}, except that they apply their method to return forecasting, whilst we are concerned with volatility forecasting.

\section{Background And Related Work}
The volatility forecasting problem is, at its core, a regression problem and there are various methods that can be used to model the data and make forecasts.

The objective of a forecasting model for volatility prediction using nonlinear regression techniques is to form a relationship of the following form:
\noindent
\begin{equation}
y = f(\mathbf{x}_n)
\end{equation}
where $\mathbf{x}_n = \left(x_1, \dots, x_n\right)$ is an input vector and $y$ is the output value (the volatility). In our problem, the previous volatility and returns are used as inputs.

The function $f$ is found by using training data to select the regression model's parameters to minimize empirical loss between the model outputs and the actual outputs. Commonly, empirical loss is defined by the sum of squared errors. For, in an ordinary least squares regression problem with a single predictor, the function $f=\mathbf{w}'\mathbf{x}$ is linear, and the fitting problem is defined as:
\noindent
\begin{equation}
min_{w} \sum_{i=n}^{n} \left( y_i - w_i x_i \right)^2
\end{equation}
where $i$ is an index variable for the data in the training sample, which has size $n$.

\subsection{SVR}
The objective in Support Vector Regression (SVR) is to minimize the size of the coefficient vector (measured by its $\ell_2$-norm), whilst requiring that the predictive accuracy of the model (measured by its $\ell_1$-norm) is at most $\epsilon$. Compared to OLS, the prediction error of the model is thus treated like a constraint. We can tune the maximum error allowable $\epsilon$ to obtain accuracy desired. The constraints and objective function thus become:
\\
\\
Minimize:
\noindent
\begin{equation}
\frac{1}{2} \left| |\mathbf{w}| \right|^2
\end{equation}
such that:
\noindent
\begin{equation}
|y_i - w_i x_i| \leq \epsilon
\end{equation}
where $i = 1,\dots,n$ represents the index for each datapoint in training data. 

It is possible that for various pairs $(x_i, y_i)$ there is no solution $w_i$ that ensures the prediction error is less than $\epsilon$. Thus, to ensure the feasibility of the optimization problem, slack variables $\xi_i, \xi_i^{*}$ can be included in the problem specification in the following manner \cite{vapnik1999nature}:

Minimize:
\noindent
\begin{equation}
\frac{1}{2} \left| |\mathbf{w}| \right|^2 + C\sum_{i=1}^{n}\left(\xi_i+\xi_i^{*}\right)
\end{equation}
such that:
\noindent
\begin{align*}
y_i - w_i x_i &\leq \epsilon + \xi_i \\
w_i x_i - y_i &\leq \epsilon + \xi_i^{*} \\
\xi_i, \xi_i^{*} &\geq 0.
\end{align*}
Here, $C$ is a hyper-parameter that controls the trade-off between the size of the coefficient vector and the tolerance of errors larger than $\epsilon$.

\subsection{Long Short Term Memory Neural Networks}
LSTMs are an enhanced version of the Recurrent Neural Network (RNN) architecture and were created to better model long-range dependencies in sequential data.

\begin{figure}
\includegraphics[width=\textwidth]{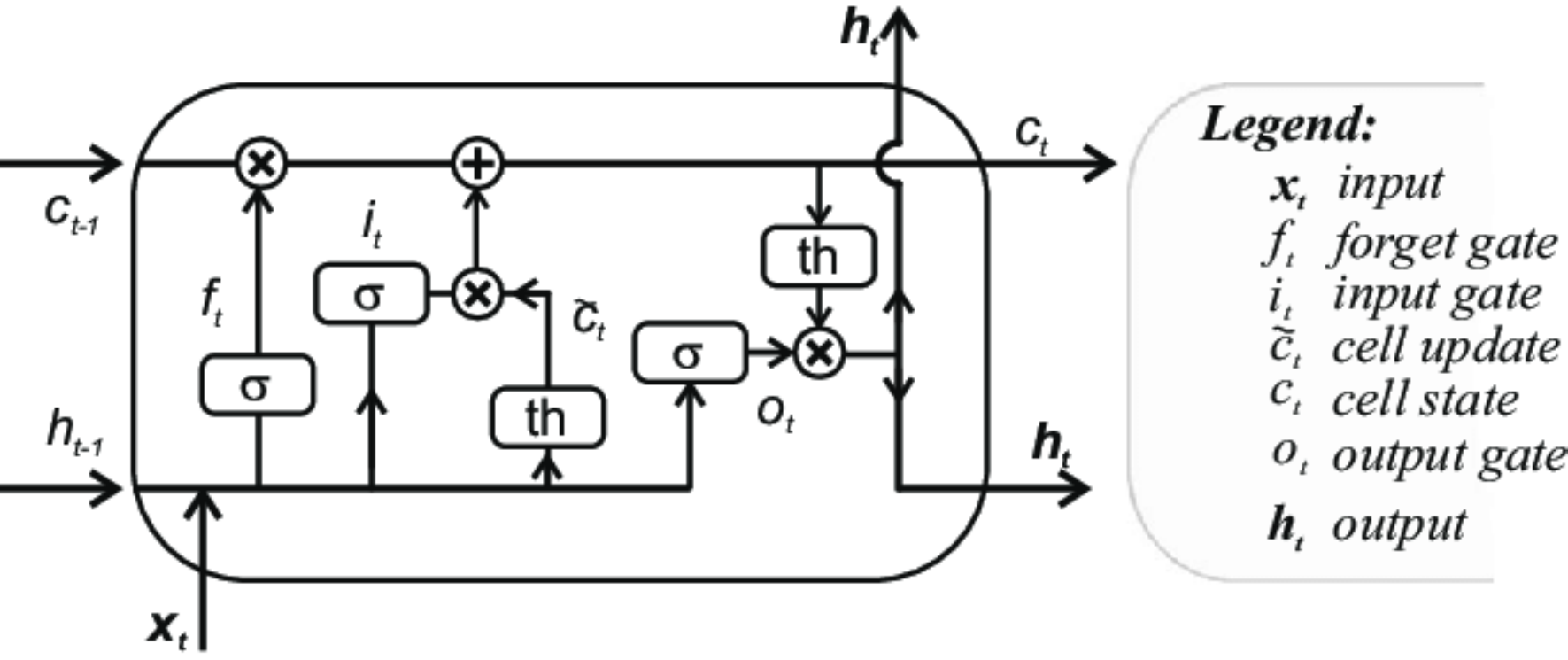}
\caption{LSTM cell \cite{hrnjica2019lake}} \label{fig1}
\end{figure}

The input at a given step along with the cell state and hidden state are the factors that affect the model's output at any given point in time. It combines these components through a series of "gates" and determines what information should be outputted and kept by the model for future use. The architecture of this model can be seen in Figure~\ref{fig1}.

The model uses three gates termed the input, forget and output gates respectively and they are used by the network as follows: 

\subsubsection{Input gate:}
The goal of this gate is to determine what new data should be added to the network's long-term memory based on the new input data and the hidden state of the previous time step. It accomplishes this through the use of a new memory and input network, both of which are neural networks.

In the new memory network, a $tanh$ activation function is used to learn how to merge the new input with the previous time step's hidden state to generate a new memory output. For the input network, a sigmoid activation function determines which parts of the new memory output are to be kept and which are to be discarded.

\subsubsection{Forget gate:}
This gate determines which information from the new input data and the previous time step's hidden state should be kept. This is accomplished by generating a vector (using the sigmoid activation function) with each element lying within the interval of $[0,1]$. Values close to 1 indicate that we want to keep that particular piece of information and values close to 0 indicate that we want to forget the information.

\subsubsection{Output gate:}
The value of the following hidden state is decided by the output gate. Information about prior inputs is contained in this state. A $sigmoid$ activation function receives the values of the current state and the prior hidden state and then a $tanh$ activation function is applied to this to give the new cell state of the system.

\subsection{GARCH(p,q)}
The GARCH(p,q) model is a traditional statistical model for modeling conditional volatility and is widely used in econometric applications. Furthermore, it is a seminal extension to the ARCH(q) model \cite{engle1982autoregressive}, which models conditional volatility as an auto-regressive process. Specifically, ARCH(q) model is given by:
\begin{align*}
\varepsilon &= \sigma_t z_t\\   
\sigma^2_t &= \alpha_0 + \sum_{i=1}^{q} \sigma_i \varepsilon^2_{t-i}
\end{align*}
where $\varepsilon$ is the deviation of the process from its mean, $z_t$ is a white-noise process with variance equal to $1$ and is independent of $\varepsilon$. The long-term variance of the process is represented by $\alpha_0$, and the variable $q$ is the autoregressive order of the process.

The GARCH(p,q) \cite{bollerslev1986generalized} model extends the ARCH(q) model by including a moving average component in the conditional volatility model:
\noindent
\begin{equation}
\sigma^2_t = \alpha_0 + \sum_{i=1}^{q} \alpha_i \varepsilon^2_{t-i} + \sum_{j=1}^{p} \beta_j \sigma_{t-i}^2.
\end{equation}

 Accordingly, setting $p=1$ and $q=1$ gives the GARCH(1,1) model. Unknown parameters are estimated through maximum likelihood estimation.

\subsection{A Hybrid Model Using Sentiment Analysis and Deep Learning}

\begin{figure}
\includegraphics[width=\textwidth]{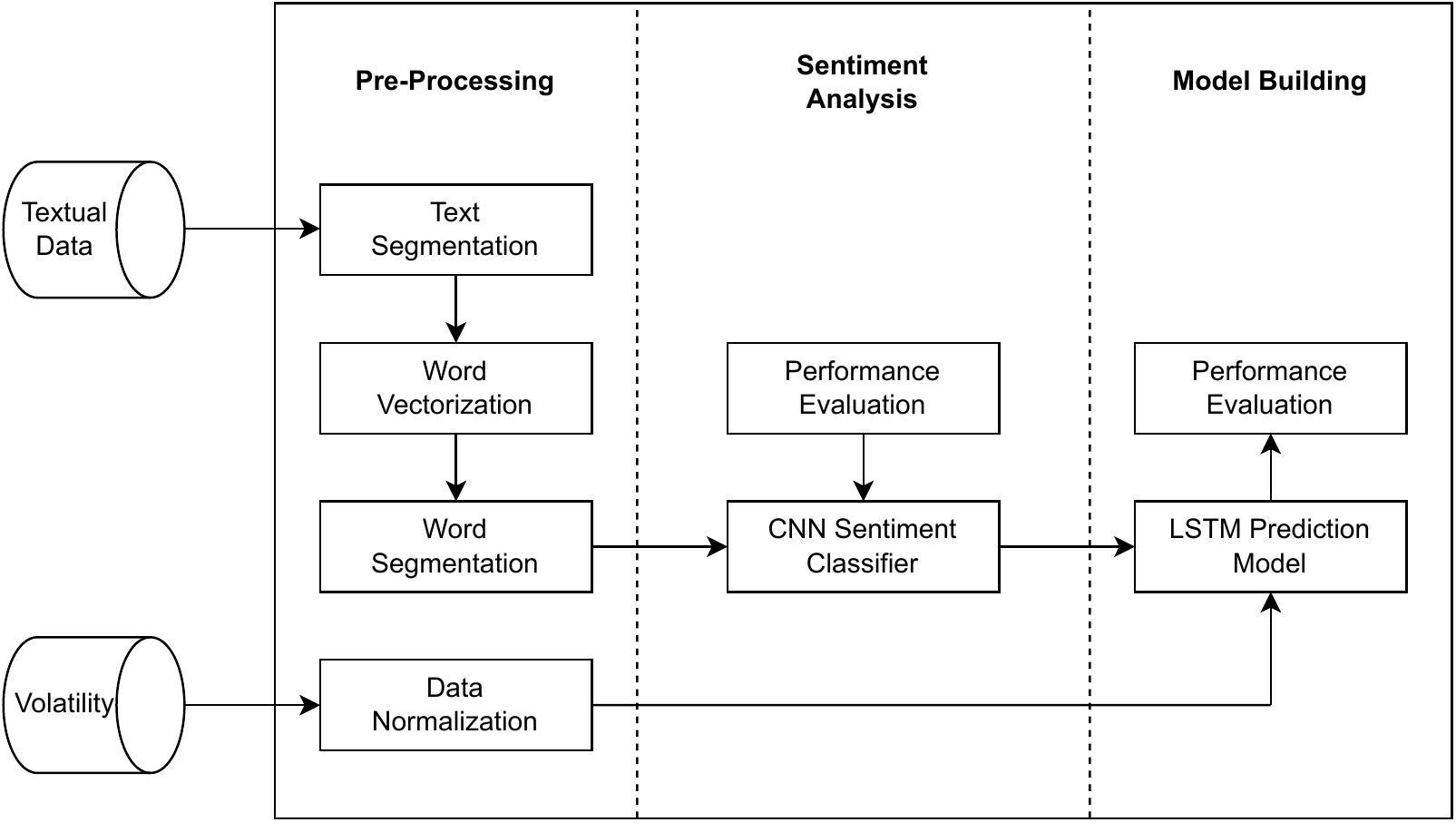}
\caption{The algorithmic workflow} \label{fig2}
\end{figure}

We propose a model that merges the sentiment extracted from Reddit global news headlines with the previous time step's volatility to predict the volatility of market indices. Our hybrid model consists of two neural networks: a CNN to perform the sentiment analysis, and an LSTM to forecast volatility. Data needs to be pre-processed (see Figure \ref{fig2}) prior to training the model, which we discuss in further detail below. 

Whilst our model is not restricted to specific hyper-parameter choices, we note that using a decreased number of layers with a large number of filters can lead us to more accurate classifications for textual data according to \cite{lee2018sentiment}. Additionally, for classification tasks, we use the F-score as the evaluation metric, but other metrics can be used. We do not explore this further.

\subsubsection{Pre-processing the textual data:}
We pre-process the textual data to filter out unnecessary noise and to transform it into something that sentiment classifiers can understand. 

Firstly, we tokenize the textual data into individual words. Secondly, we remove all the stop words: a list of commonly used words in a language that don't contribute to the sentiment of a piece of text. We do this to increase the efficiency of the classifying model and for dimensionality reduction purposes.

We utilize word2vec, a natural language processing tool published by Google in 2013 to learn the interdependence of words in a corpus. It uses the Skip-gram model and the continuous bag of words model to output a representation of the words in vector form. The Skip-gram model uses the central word to predict the context of a piece of text, and the continuous bag of words architecture uses the context to predict the central word.

When developing the model, we use a global news headlines corpus obtained from Reddit for training which contains 27 headlines per trading day. Given that there is $n$ number of words in the text feature of a sentence, a vector embedding of length 100 to 800 is obtained after utilizing word2vec as mentioned by \cite{mikolov2013efficient}.

\section{Methodology}
We evaluate our proposed model on daily historical data of several stock market indices against the following benchmark methods: GARCH(1,1), SVR, and an LSTM without sentiment data. The historical data is sourced from the Wall Street Journal and covers the period 8 August 2008 to 1 July 2016, but the amount of data differs slightly across indices (see Table \ref{tab:data}). To facilitate simpler implementation, volatility is approximated as the squared log return. 

The historical data was first split into training and test sets. The training data was used to estimate model parameters and for hyper-parameter tuning. In particular, hyper-parameter tuning was performed for the SVR model using 20-fold cross-validation. 

To evaluate the models, the Root Mean Square Error is calculated on the test set. Furthermore, we test if the models' conditional volatility forecasts are superior to a constant volatility estimate. To do this, we regress the actual data onto the predictions and evaluate if the predictions are significantly superior to the mean using the F-test. 

\begin{table}[htb!]
\caption{The Data}
\begin{center}
\begin{tabular}{l|rr}
Index & \textbf{\makecell[cr]Training period (size)} & \textbf{\makecell[cr]{Testing period (size)}}\\
\bottomrule\toprule
S\&P500 & 08/08/2008-2/12/2014 (1359) & 3/12/2014-01/07/2016 (631) \\
Ibovespa & 08/08/2008-2/12/2014 (1337) & 3/12/2014-01/07/2016 (618) \\
RTS-50 & 08/08/2008-2/12/2014 (1406) & 3/12/2014-01/07/2016 (652) \\
Nifty-50 & 31/12/2008-2/12/2014 (1238) & 3/12/2014-01/07/2016 (613) \\
SCHOMP & 08/08/2008-2/12/2014 (1310) & 3/12/2014-01/07/2016 (610) \\
JSE top 40 & 08/08/2008-2/12/2014 (1428) & 3/12/2014-01/07/2016 (630) \\
\bottomrule
\end{tabular}
\label{tab:data}
\end{center}
\end{table}

The GARCH model's parameters were pre-specified as $p=1$ and $q=1$. For Support Vector Regression, the optimal parameters obtained using a grid search was the Radial Basis Function (RBF) kernel with $\gamma=0.001$ and $C=2$.

The CNN-based sentiment model was trained with 100-dimensional word2vec embeddings derived from the Reddit global news headlines corpus (headlines made available for trading days), had 128 filters, one global max pooling layer, and the sigmoid function as the activation function in the output layer. The sentiment model was then benchmarked against a Random Forest and a Logistic Regression model.

Sentiment predictions were then fed as input to the LSTM sentiment-based model to make the final volatility forecast. Our LSTMs were implemented using the Keras library and used the previous day's volatility as the input. Our proposed method, the sentiment LSTM, received sentiment as an additional input variable. The parameters for the LSTM were a dropout rate of $0.2$, the output layer a dense layer with 1 unit, and 30 neurons in the hidden layer.

Additionally, we shifted the sentiment predictions by one day to examine the extent to which the information in the next day's sentiment, which is not present in the current day's sentiment, can improve the volatility forecast for the next day. Thus, strictly speaking, this is not a forecast, but rather serves to investigate the explanatory information present in sentiment. This means that instead of the LSTM model receiving the previous time step's volatility and sentiment, we fed it the previous time step's volatility and the current time step's sentiment.

\section{Results And Discussion}

\begin{table}[htb!]
\caption{Test RMSE on the S\&P 500}
\begin{center}
\begin{tabular}{l|rr}
Predictive model & \textbf{\makecell[cr]RMSE} & \textbf{\makecell[cr]{p-value}}\\
\bottomrule\toprule
GARCH(1,1)                  & $9.86\cdot10^{-03}$           & $\approx0 < 0.05$ \\
SVR                         & $\mathbf{1.83\cdot10^{-04}}$  & $\approx0 < 0.05$ \\
LSTM                        & $1.94\cdot10^{-04}$           & $\approx0 < 0.05$ \\
LSTM with sentiment         & $2.00\cdot10^{-04}$           & $\approx0 < 0.05$ \\
LSTM with sentiment shifted & $1.92\cdot10^{-04}$           & $\approx0 < 0.05$ \\
\bottomrule
\end{tabular}
\label{tab2}
\end{center}
\end{table}

The Support Vector Regression model was the best performing model on the S\&P500 whilst the GARCH(1,1) was the best performing model on the Nifty-50, as shown in Tables \ref{tab2} and \ref{tab3} respectively. We can also observe that all models obtained statistical significance with a p-value close to zero indicating that all models were superior to a constant variance. 

\begin{table}[htb!]
\caption{Test RMSE on the Ibovespa}
\begin{center}
\begin{tabular}{l|rr}
Predictive model & \textbf{\makecell[cr]RMSE} & \textbf{\makecell[cr]{p-value}}\\
\bottomrule\toprule
GARCH(1,1)                  & $1.20\cdot10^{-06}$          & $\approx0 < 0.05$ \\
SVR                         & $1.33\cdot10^{-06}$          & $\approx0 < 0.05$ \\
LSTM                        & $1.26\cdot10^{-06}$          & $\approx0 < 0.05$ \\
LSTM with sentiment         & $\mathbf{1.17\cdot10^{-06}}$ & $\approx0 < 0.05$ \\
LSTM with sentiment shifted & $1.20\cdot10^{-06}$          & $\approx0 < 0.05$ \\
\bottomrule
\end{tabular}
\label{tab3}
\end{center}
\end{table}

\begin{table}[htb!]
\caption{Test RMSE on the Nifty-50}
\begin{center}
\begin{tabular}{l|rr}
Predictive model & \textbf{\makecell[cr]RMSE} & \textbf{\makecell[cr]{p-value}}\\
\bottomrule\toprule
GARCH(1,1)                  & $\mathbf{1.74\cdot10^{-07}}$ & $\approx0 < 0.05$ \\
SVR                         & $2.59\cdot10^{-07}$ & $\approx0 < 0.05$ \\
LSTM                        & $2.05\cdot10^{-07}$ & $\approx0 < 0.05$ \\
LSTM with sentiment         & $1.85\cdot10^{-07}$ & $\approx0 < 0.05$ \\
LSTM with sentiment shifted & $1.99\cdot10^{-07}$ & $\approx0 < 0.05$ \\
\bottomrule
\end{tabular}
\label{tab5}
\end{center}
\end{table}

\begin{table}[htb!]
\caption{Test RMSE on the SHCOMP}
\begin{center}
\begin{tabular}{l|rr}
Predictive model & \textbf{\makecell[cr]RMSE} & \textbf{\makecell[cr]{p-value}}\\
\bottomrule\toprule
GARCH(1,1)                  & $1.02\cdot10^{-05}$ & $\approx0 < 0.05$ \\
SVR                         & $7.64\cdot10^{-06}$ & $\approx0 < 0.05$ \\
LSTM                        & $6.92\cdot10^{-06}$ & $\approx0 < 0.05$ \\
LSTM with sentiment         & $\mathbf{6.25\cdot10^{-06}}$ & $\approx0 < 0.05$ \\
LSTM with sentiment shifted & $6.45\cdot10^{-06}$ & $\approx0 < 0.05$ \\
\bottomrule
\end{tabular}
\label{tab6}
\end{center}
\end{table}

\begin{table}[htb!]
\caption{Test RMSE on the RTS-50}
\begin{center}
\begin{tabular}{l|rr}
Predictive model & \textbf{\makecell[cr]RMSE} & \textbf{\makecell[cr]{p-value}}\\
\bottomrule\toprule
GARCH(1,1)                  & $2.80\cdot10^{-05}$ & $\approx0 < 0.05$ \\
SVR                         & $2.10\cdot10^{-05}$ & $\approx0 < 0.05$ \\
LSTM                        & $1.82\cdot10^{-05}$ & $\approx0 < 0.05$ \\
LSTM with sentiment         & $1.84\cdot10^{-05}$ & $\approx0 < 0.05$ \\
LSTM with sentiment shifted & $\mathbf{1.76\cdot10^{-05}}$ & $\approx0 < 0.05$ \\
\bottomrule
\end{tabular}
\label{tab4}
\end{center}
\end{table}

\begin{table}[htb!]
\caption{Test RMSE on the JSE top 40}
\begin{center}
\begin{tabular}{l|rr}
Predictive model & \textbf{\makecell[cr]RMSE} & \textbf{\makecell[cr]{p-value}}\\
\bottomrule\toprule
GARCH(1,1)                  & $2.67\cdot10^{-07}$ & $\approx0 < 0.05$ \\
SVR                         & $3.11\cdot10^{-07}$ & $\approx0 < 0.05$ \\
LSTM                        & $2.73\cdot10^{-07}$ & $\approx0 < 0.05$ \\
LSTM with sentiment         & $2.75\cdot10^{-07}$ & $\approx0 < 0.05$ \\
LSTM with sentiment shifted & $\mathbf{2.66\cdot10^{-07}}$ & $\approx0 < 0.05$ \\
\bottomrule
\end{tabular}
\label{tab7}
\end{center}
\end{table}

\begin{figure}[htb!]
\includegraphics[width=\textwidth]{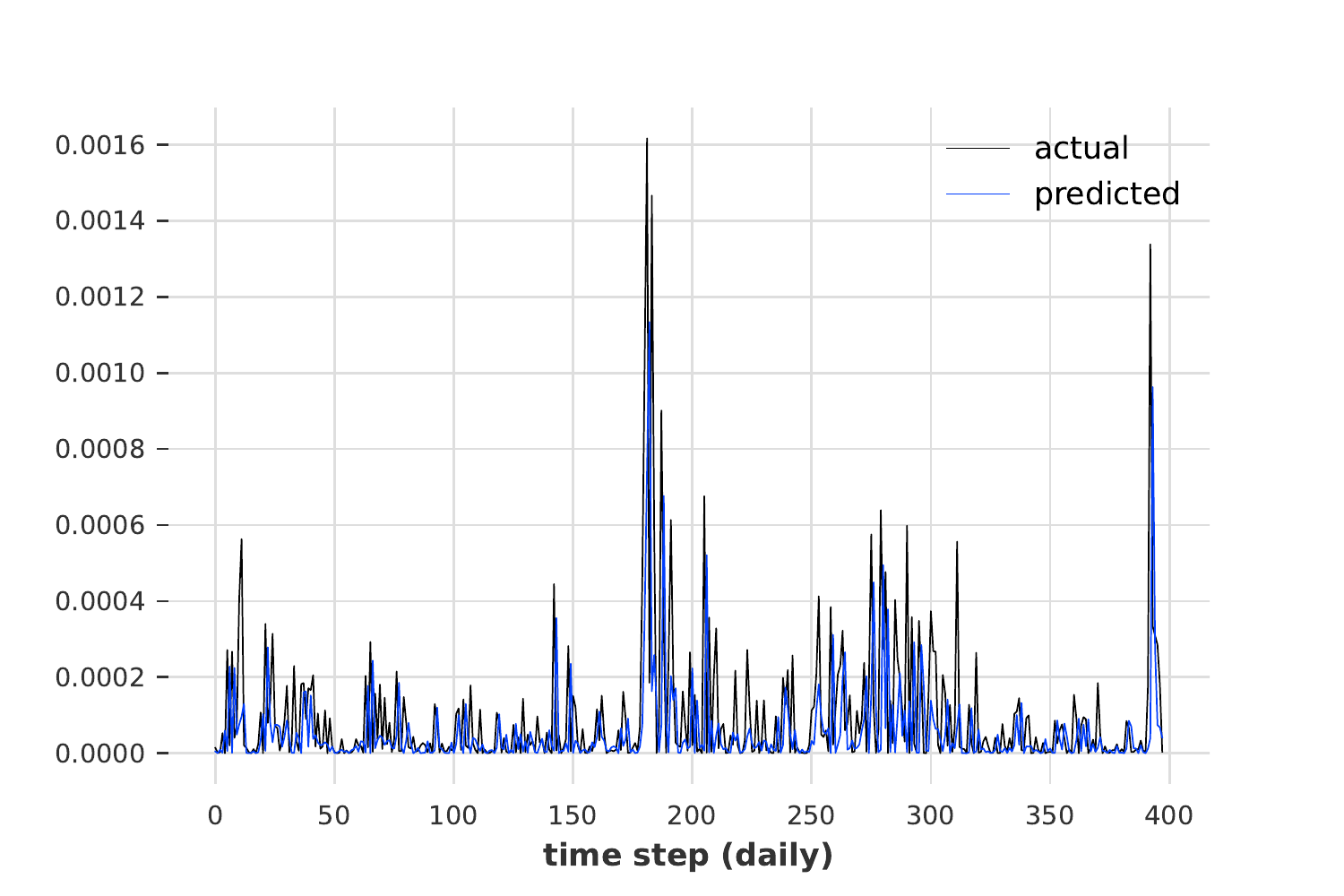}
\caption{LSTM with sentiment predictions on the S\&P500 market volatility squared.} \label{fig3}
\end{figure}

\begin{figure}[htb!]
\includegraphics[width=\textwidth]{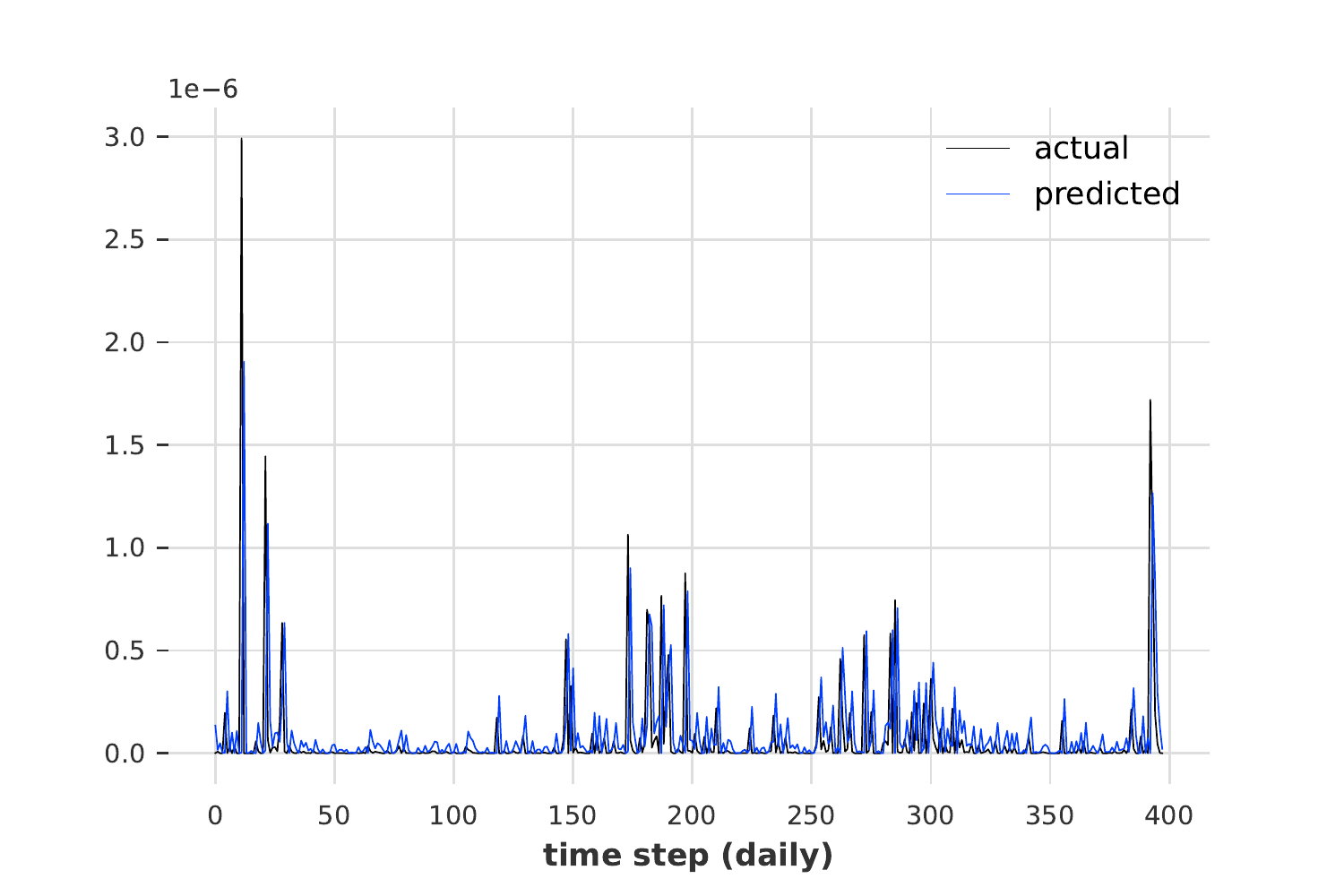}
\caption{LSTM with sentiment predictions on the JSE top 40 volatility squared.} \label{fig4}
\end{figure}

Surprisingly, we find that the LSTM with shifted sentiment was the best performer for two markets only, the RTS-50 and JSE Top 40 (Tables \ref{tab4} and \ref{tab7} respectively), corresponding with Russian and South African markets. In contrast, it appears that the previous day's sentiment data provided better volatility forecasts on the Ibovespa and SHCOMP datasets (Table \ref{tab3} and \ref{tab6} respectively), corresponding Brazilian and Chinese markets.

In Figures \ref{fig3} and \ref{fig4}, we can see the plot of our method's predictions on the S\&P500 and JSE top 40. The plots show the actual and predicted volatility.

\begin{table}[htb!]
\caption{Results of the sentiment classifier}
\begin{center}
\begin{tabular}{l|rrr}
Metric & \textbf{CNN} & 
\textbf{\makecell[cr]{Logistic\\Regression}} & 
\textbf{\makecell[cr]{Random\\Forest}}\\
\bottomrule\toprule
precision & 0.85          & 0.84 & \textbf{0.89} \\
recall    & \textbf{0.87} & 0.84 & 0.85 \\
F-score   & \textbf{0.86} & 0.84 & 0.85 \\
\bottomrule
\end{tabular}
\label{tab1}
\end{center}
\end{table}

Lastly, the Convolutional Neural Network showed better results for sentiment classification - outperforming the benchmark classifiers of Logistic Regression and Random Forest (Table \ref{tab1}) with an F-score of 0.86.

\subsection{Discussion Summary}
Notably, our results show that there is no clear outperforming method across all markets. Notably, in two markets, we are presented with the result that future sentiment is less predictive of future volatility than current sentiment. If Reddit posts were commenting on the market conditions for the same day, then we would expect that future sentiment should be a superior predictor of volatility. However, our results seem to indicate that this is not necessarily the case.

Furthermore, the findings of our research do not imply that our model will perform in the same manner on individual financial assets. Individual financial assets have additional sources of variation due to idiosyncratic risks. Thus, textual data for sentiment analysis would have to be more specific than global news headlines. This finding highlights the need for more research on sentiment as an input for predicting individual financial assets as most studies, including~\cite{jing2021hybrid} and \cite{mehtab2019robust}, used financial indices.

In developing our model, we have also confirmed the findings of \cite{al2018deep} and \cite{chen2020verbal} who proved that neural networks are more successful than traditional machine learning approaches at text classification tasks. 

\section{Conclusion}
In this paper, we used an LSTM with sentiment input from a CNN to forecast the volatility which we evaluated on historical data of the S\&P500, Ibovespa, RTS-50, Nifty-50, SHCOMP, and JSE top 40 indices. We used a CNN to extract public sentiment from Reddit global news headlines data as described by \cite{jing2021hybrid} and observed that our model showed better results than the other benchmark classifiers.

Our volatility forecasting results demonstrated that sentiment input can add predictive power to a volatility forecasting model, but this appears to be market specific.  Although the LSTM with sentiment did not outperform the benchmarks in some markets, it did provide more accurate forecasts than the LSTM without sentiment input. 

Furthermore, we shifted the sentiment predictions to feed the LSTM model the present step's sentiment forecast to examine if there is a contemporaneous correlation between sentiment and volatility. This was done to examine how much more information was contained in current sentiment than sentiment from the previous day. Surprisingly, we found that incorporating future sentiment did not always increase the accuracy of the volatility forecast beyond incorporating current sentiment. However, it should be noted that our results may be contaminated by the variation of the expected return, which for simplicity, was not explicitly modeled.

\bibliographystyle{splncs04nat}
\bibliography{ref}

\end{document}